# DIFFERENCE OF PROBABILITY AND INFORMATION ENTROPY FOR SKILLS CLASSIFICATION AND PREDICTION IN STUDENT LEARNING

Kennedy Efosa Ehimwenma[1], Safiya Al Sharji[2] and Maruf Raheem[3]

[1]Department of Computer Science, College of Science and Technology,
Wenzhou-Kean University, China
[2]Department of Information Technology,
University of Technology and Applied Sciences, Muscat, Oman
[3]Department of Statistics University of Uyo, Nigeria

## ABSTRACT

*The probability of an event is in the range of [0, 1]. In a sample space S, the value of probability determines whether an outcome is true or false. The probability of an event Pr(A) that will never occur = 0. The probability of the event Pr(B) that will certainly occur = 1. This makes both events A and B thus a certainty. Furthermore, the sum of probabilities Pr(E1) + Pr(E2) + ... + Pr(En) of a finite set of events in a given sample space S = 1. Conversely, the difference of the sum of two probabilities that will certainly occur is 0. Firstly, this paper discusses Bayes' theorem, then complement of probability and the difference of probability for occurrences of learning-events, before applying these in the prediction of learning objects in student learning. Given the sum total of 1; to make recommendation for student learning, this paper submits that the difference of argMaxPr(S) and probability of student-performance quantifies the weight of learning objects for students. Using a dataset of skill-set, the computational procedure demonstrates: i) the probability of skill-set events that has occurred that would lead to higher level learning; ii) the probability of the events that has not occurred that requires subject-matter relearning; iii) accuracy of decision tree in the prediction of student performance into class labels; and iv) information entropy about skill-set data and its implication on student cognitive performance and recommendation of learning [1].*

## 1. INTRODUCTION

Indeed, the theory of probability application is far beyond determining the outcome of a rolled dice, picking a colored ball from an urn or a card from a set in a deck of cards [2]. Probability is thus a veritable method for drawing inferences i.e. arriving at conclusions from given evidence and reasoning in a given domain problem. While approaches such as deductive and inductive logic are based on logical reasoning from available evidence, approaches by probabilistic methods are largely dependent on statistical computation of facts and the evidence provided.

Assessment is an indispensable part of teaching and learning process. For subject-matter mastery, this paper argues using an adaptable predictive algorithmic model that pre-learning assessment which unravels and determines a student's state of knowledge before teaching and further learning should be an integral part of teaching and learning systems. That, this should be achieve by building ML algorithm into teaching, learning and assessment (LTA) systems to become (pre-assessment) PLTA systems.

This paper first discusses Bayes, complement of probability or difference of probability for occurrences of learning events, before applying them in the predication of learning objects in student learning. Like Bayes, the difference of probability computation is a model of supervised learning algorithm for recommender systems. Within a sample space $S$, the total probability $argMaxPr(S) = 1$. Probability is a measure of the ratio of an event $E$ in comparison to total number of occurrences $S$. In a finite space $A$, the complement of probability of an event denoted by $Pr(\neg E)$, is that which is not occurring; and of that which has occurred is given as $Pr(E)$. Therefore

$$Pr(E) + Pr(\neg E) = 1 \quad (1)$$

$$Pr(\neg E) = 1 - Pr(E) \quad (2)$$

On the basis of Bayes, probability is a supervised learning technique for intelligent prediction, classification and recommendation of actions to be taken as per effective decision making. Supervised machine learning technique takes a set of inputs $x_i$ and outputs $y_i$ as training model $T(x_i, y_i)$. This is in turn used for future prediction denoted as $T(x_i, ?)$ based on a set of unknown but related dataset present in the training set; where the wildcard ? is the expected prediction. In this paper, we show that the computation of the $Pr(\neg E)$ on the basis of complement theorem is the student missing skill-set. Given a set of $1, 2, 3, ..., n$ learning objects where $1 \leq i \leq j \leq k \leq n$, according to Bayes' rule e.g. [3, 4, 5]; it holds that the $Pr(\neg E)$ is

$$Pr(E_i|A) = \frac{Pr(E_i \cap A)}{\sum_{i=1}^{n} Pr(A \cap E_i)} \quad (3)$$

which determines the *maximum A posterior* probability value of the pre-assessment object $E_i$ amongst the $n$ learning modules in the sample space $A$ or objects that are recommended to students after some pre-assessments performance.

### 1.1. Contribution of this Paper

The contributions of this paper are

- *i)* the computation of probability of the occurrence of student skill-set,
- *ii)* complement of probability computation as a representation of skill-sets that are required by students (because the skills did not occur in the student knowledge given that they were failed at pre-assessment),
- *iii)* verification and applicable extension of the difference of probability computation in students' pre-skill assessments using Bayes' rule, and
- *iv)* estimating *uncertainty* with student performance dataset and the effect of *uncertain* prediction on learning material.

The rest of this paper continues with Section 2 presenting related works on the theory of probability. Section 3 presents the difference of probability and its relationship with complement of probability, and in furtherance to Bayes' theorem for prediction in student learning. With the use of the dataset gathered from prior study on the pre-assessments of SQL query statements, the section showed the application of Bayes' in the computation of recommendation of learning objects. From our programmable architecture, section 4 discusses our programmable calculation and the iterative process involved in the derivation of the predictive probability theory for recommendation of learning. In addition, the section calculated information entropy that determines the impurity in the skill-set data as well as decision tree modelling from the dataset. Section 5 conclusions.

## 2. RELATED WORKS

By and large, inference probability approaches are applied to a range of applications. With emphasis on binary classification such as defective or non defective, the work of [6] surveyed several application areas and sub-areas in which binary function $q \epsilon \{0,1\}$ classification have been used. The areas covered in the survey are namely; medical testing and disease classification/treatment (via rule based classification method for clinicians using AND or OR clauses), biology (DNA testing, Counting defective items), telecommunications (Multiple access channels, Cognitive radios), information technology (Data storage and compression, cyber security, database systems, Bloom filters), and data science (Search problems, Sparse inference and learning, Theoretical computer science). Probability in such areas as these are applied to discover small entities within a large pool of items [6].

Mathematical models of classification and prediction are generally geared towards the reduction of human intervention in our daily routines and process and to improve efficiency and productivity. In mental health evaluation of college students, [7] observed that through the use of fuzzy mathematics in combination with entropy weight on analysis of experimental data, comprehensive mathematical evaluation model is established. In their research study, [8] also applied the entropy weighted method and Discrete Hopfield neural nets for the evaluation and classification of higher education systems and standards in different countries using MATLAB. As such, there are many approaches to the problem of classification and prediction. In table 1, this paper presents a brief summary of four computational models for decision making in classification and their characteristics; and in furtherance Bayes' theorem which is the basis for this paper.

Table 1: Comparison of Some of Computational Decision Models [9]

| Model | Comparison | Performance Metric | Recommendation |
|---|---|---|---|
| Naïve Bayes | *Naïve Bayes is popular and suitable for large input data. *Each feature in the input vector are conditionally independent. *This ML technique has low computational(space and time) complexity. | - | *Probability computation based on the application of Bayes formula. *For large dataset. |
| CART (classification and Regression Tree) | *Ability to generate fuzzy rules for prediction purposes. | - | *For rule-based prediction and generation. |
| Random Forest | *Has the ability to perform better than decision tree algorithms. | - | *For aggregation of $n$ number of decision trees. |
| J48 (Decision Tree algorithm) | *J48 is a ML algorithm that has the ability to select specific features or instances and theirmissing attributes. *It has the ability to support both continuous and categorical instances in the process of tree construction. | J48 (67.15%) had better performance than CART (62.28%), SVM (65.04%), KNN (53.39%). | *High prediction accuracy. *Easy to use. *Tree pruning. *Classification of tree into positive and negative instances, respectively. |

## 2.1. Bayes

Naive Bayes has been ascertained to be a simple algorithm, yet performs very well [10] e.g. in text and document analysis and classification [11] [12] [13] [14] [15]. In [5], when the "prior" probabilities $P(A_i)$ and the likelihood $P(B|A_i)$ to obtain B for each $A_i$ are known for a number of events $A_i = E\ space$, then it holds that

$$P(B|A_i) = \frac{P(B|A_i).\ P(A_i)}{\sum_{i=1}^{n} P(B|A_j) P(A_j)}.$$

Bayes theorem or rules is a machine learning technique for classifying and making predictive analysis after learning from a set of data and events for effective decision making [16]. On classification problem, the work of [16] reported and argued that studies have observed that Bayes algorithm has better performance, e.g. [17, 18]. J48, on the other hand, has also been shown to have better performance given the work of [19] . In the comparison of Naïve Bayes to Complement Naïve Bayes, [16] states that both are algorithms that are commonly used for texts classification because of their fast and easy implementation. In the same vein, that both has shown some advantages and disadvantages. From experimentation and result, [16] observed that performance of Complement Naïve Bayes is better than performance of Naïve Bayes.

## 2.2. The Objective and Subjective Approach to Prediction

The foundation of probability theory has been widely attributed to a game of chance experiment [3] [5] [20]. Yet, further development has seen increased approaches to probability application in different domains of machine learning. In learning, teaching, and assessment (LTA), "*knowing*" is subjective to the "*knower*". The question is: can probability be applied in: *i)* prediction of the outcome of a student's (pre)assessment, and *ii)* prediction of the requisite knowledge: on the basis of the skill-sets that has been *learned*; *not-yet-learned* or *learned-and-unlearned* (i.e. forgotten)? Conversely, is the *law of large number* [5] [20] adequate enough to provide answers to this questions? Data obtained from experiment are always objective [5]. But in LTA, the probability of predicting a student to learn some *failed* skill-set or to progress to a higher learning after all pre-assessments are *passed* is subjective to the dataset at hand and to a specific domain of learning content. This is just as the knowledge of the student on a given content is subjective to that content or student. Hence, the more reason why probability has been described as a personal degree of belief and ability that depends on a person's knowledge, experience or specific context [5]. A concept can be learned and unlearned and then re-learned. This is because learning can be fluid and highly volatile.

## 2.3. The Expected Value

The computation of all probabilities of the events or elements in a sample space $A = 1$ which is equivalent to 100%. In teaching and learning, the goal of any student is to achieve 100% score performance during evaluation. This is the *expected maximum value*. A student attaining the probability of 1 in a given pre-assessment episode must have *passed* all the pre-assessment quizzes, and in that manner, progresses to a higher level of learning — called the student's *desired* concept [21] [22] [23] [24]. Thus, the expected maximum value for a set of all *passed* performance $P_i$ where $1 \leq i \leq j \leq k \leq n$ is defined as

$$argMaxPr(P_i) = 1 \quad (4)$$

In [20] this is stated as

$$\sum_{\omega \in E} m(\omega) = 1$$

where event $\omega \in \Omega$ space, and $m$ a real-value distribution function such that $m(\omega) \geq 0$. Therefore, the **maximum A posterior** in *eq.3* and *eq.4* are equivalent and are defined as

$$\sum_{i=1}^{n} Pr(Pass_i | A) \equiv \text{argMax} Pr(Pass_i | A) = 1 \quad (5)$$

Then, it holds that

$$Pr(Fail_k) = \sum_{i=1}^{n} Pr(Pass_i | A) - \sum_{j=1}^{n} Pr(Pass_j | A) \quad (6)$$

where $Pr(Fail_k)$ is the probability weight of the $k$ learning objects needed by the student. From the foregoing, it also holds that the difference of probability equals Bayes' probability output which also predicts the needed learning object. Thus, combining *eq. 3* and *eq. 6*, we have

$$Pr(E_i | A) = \sum_{i=1}^{n} Pr(Pass_i | A) - \sum_{j=1}^{n} Pr(Pass_j | A) \quad (7)$$

## 3. METHOD OF DIFFERENCE OF PROBABILITY FOR STUDENT LEARNING PREDICTION

In this work, the difference of probability computation is based on the complement theorem of sets. It is a technique that is employed to compute the probability that represents appropriate materials needed for student learning based on some pre-assessment dataset in [23]. In formal education, learning is sequential and chronological given the order of learning in curricula. In Figure 1 is a knowledge graph that represents a set of knowledge modules in SQL programming which are interdependent on each other in accordance to the structure. The graph comprises three parent nodes, namely, *delete*, *insert* and *select* that are linked to their respective leaf nodes. The inter-relationship between parent nodes depicts that, for instance, the *select* node is a prerequisite to both *insert* and *delete* nodes. In other words, to learn the node *insert* or *delete*, a student must have a complete mastery of the leaf nodes pre-assessed upon and *passed all* the leaf nodes of the *select* module by the pre-assessment system [21] [23]. The difference of probability computation of this paper is illustrated considering the graph structure of Figure 1.

Now, let us consider the *set S* of two learning objects as

$$S = \{delete, select\} \quad (8)$$

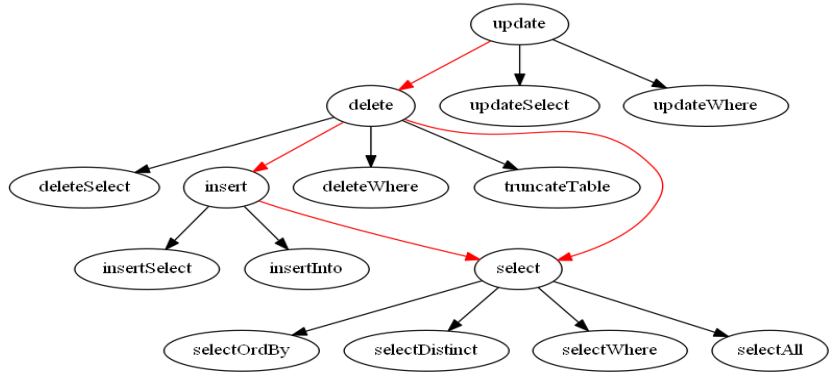

Figure 1: A knowledge graph of finite set of learning objects of multiple horizontal and vertical traversal

Based on Figure 1, the objects *delete* and *Select* are *parent nodes* to some other leafnode objects. Using the notation $S = \{\{S_1\},\{S_2\}\}$, we have

$$A = \{Delete = \{truncateTable, deleteSelect, deleteWhere\},$$
$$Select = \{selectOrderBy, selectDistinct, selectWhere, selectAll\}\} \quad (9)$$

Since, each leafnode object have two possible states *Pass* and *Fail*, then we have the *<Pass | Fail>* decision-state per leafnode for the student skills status assessments in the form

$$S = \{\{S_1\{< Pass|Fail >\}\}, \{S_2\{< Pass|Fail >\}\}, \dots, \{S_n\{< Pass|Fail >\}\}\} \quad (10)$$

From *eq.10*, we expand

**S** = {**delete** = {deleteSelect = {< $Pass|Fail$ >}, deleteWhere = {< $Pass|Fail$ >}, truncateTable = {< $Pass|Fail$ >}},

**select** = {selectOrderBy = {< $Pass|Fail$ >}, selectDistinct = {< $Pass|Fail$ >}, selectWhere ={< $Pass|Fail$ >},

selectAll = {< $Pass|Fail$ >}}} (11)

### 3.1. Bayes' Theory of Classification and Object Prediction

In this section, we present the prediction model of the needed (or recommended) learning on the basis of Bayes' rule. If a student *fails* a pre-assessment quiz, the computed probability, as shown below, is a representation of the proportion of the quiz(zes) relative to the overall *pre-assessment performance*. The computation of needed learning using Bayes', is also computable using complement theory or difference of probability as shown from *eq. 1 - 7*. Given *eq.10*, it is implied that the recommended materials are based on the set $A = < Pass\ |\ Fail >$ student performance

From Bayes', specifically, for a higher level of topic prediction, i.e. the probability of getting to learn a *Desired_Concept* given the performance of all *passed* pre-assessments $\boldsymbol{Performance_{Pass}}$ is

$$Pr(DesiredConcept\ |\ Performace_{pass}) = \frac{Pr(Performace_{pass}\ |\ DesiredConcept) * Pr(DesiredConcept)}{Pr(Performace_{pass})}$$
(12)

The probability of having a leafnode recommended given the set of performance is

$$Pr(LeafNode|\ Performance) = \frac{Pr(performance|leafnode)*Pr(leafnode)}{Pr(Performance)} \quad (13)$$

Computing probability for a *Leafnode* implies that the *Leafnode* has not been *passed*. As such, it is not yet part of the required skill-set to progress to a higher-level node learning in the knowledge graph. Essentially, $eq.12$ and $13$ generalizes to

$$Pr(Node_j|Performance) = \frac{Pr(Performance\ |\ Node_j) * Pr(Node_j)}{\sum_{i=1}^{n} Pr(Performance\ |\ Node_i) * Pr(Node_i)} \quad (14)$$

where *DesiredConcept*, *LeafNode* or *Node* are quantities which reflects the weighted probability of learning given the students' performance.

With regards to *eq.11*; suppose a learning object is chosen at random for pre-assessment, what is the probability of having that same object recommended for learning: In other words, it is having a **Fail** in that pre-assessment, therefore the probability of a **Fail** (which is similar to saying the probability of recommending the $Node_j$ in *eq.14* is

$$Pr(Fail_j|A) = \frac{Pr(Fail_j \cap A)}{\sum_{i=1}^{n} Pr(Fail_i \cap A)} \quad (15)$$

Having stated the probability of a node prediction on the basis of Bayes' rule; now using a set of student performance data from Table 2--- an extraction from the pre-assessment dataset of [23], in the following section; we then provide cases to illustrate the foregoing formulas.

### 3.2. Case I: A Single Parent-Node Object and Probability of Instance Leafnode Prediction

Let us consider the parent-node *select* (Fig. 1) where the elements of $select = \{SOB, SD, SW, SA\}$. From the dataset (Table 1), a given student's performance that is also shown in an *onto mapping* (Fig. 2) is the set

$$\boldsymbol{Performance_{student}} = \{Fail, Pass, Pass, Pass\} \quad (16)$$

Then, the *marginal probability* for the set given in *eq.16* is

$$Pr(Fail_{SOB}) \equiv Pr(SOB) = \frac{1}{4} = 0.25$$

which represents the probability weight of the needed knowledge by the student.

Table 2. Cross-section of individual-student performance

| Parent Node | Child Objects | Pass | Fail | Total |
|---|---|---|---|---|
|  | SOB | 0 | 1 | 1 |
| Select | SD | 1 | 0 | 1 |
|  | SW | 1 | 0 | 1 |
|  | SA | 1 | 0 | 1 |
|  | TT | 1 | 0 | 1 |
| Delete | DW | 0 | 1 | 1 |
|  | DS | 0 | 1 | 1 |
| Total |  | 4 | 3 | 7 |

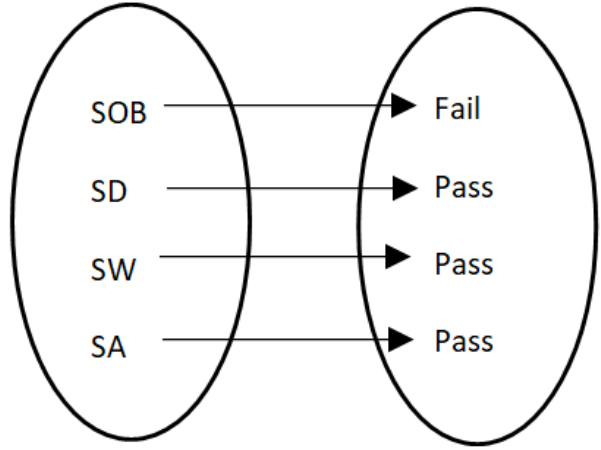

Figure 2. Onto mapping of children of the Select instance to some student performance outcome.

### 3.3. Case II: Multiple Parent-Node Objects and Probability of Instance Leafnode Recommendation

Again, from Figure 1, let the set $A = \{Select, Delete\}$; and from *eq. 9*, we have the set $A = \{\{SOB, SD, SW, SA\}, \{DS, DW, TT\}\}$. Based on Table 2, we have the set of a given *student performance* as

$$\mathbf{Performance}_{student} = \{\{Fail, Pass, Pass, Pass\}, \{Fail, Fail, Pass\}\} \quad (17)$$

Applying the rule in *eq. 15*, the probability of recommending a *failed* node for re-learning in the subset of $A = select$ parent-node is computed as follows:

$$Pr(Fail_{SOB}|A) = \frac{Pr(Fail_{SOB} \cap A)}{Pr(Fail_{SOB} \cap Select) + Pr(Fail_{Delete} \cap Delete)} \quad (18)$$

Suppose a student make a random choice of a *Desired_Concept* to learn, which maybe higher in hierarchy such as the parent-node **delete**(Figure 1). Automatically, that makes the prerequisite nodes to the $Desired\_Concept_{prereq} = \{select, insert\}$ and $Desired\_Concept = \{delete\}$. Since we have been working with the set $A = \{select, delete\}$; firstly, we need the probability of a random choice that resulted in the *student performance* of *eq.17*. Thus, from the two parent - objects $A = \{select, delete\}$, the probability of recommending a node to learn is

$$Pr(Select) = Pr(Delete) = \frac{1}{2} = 0.5$$

From *eq. 17*,

$$\mathbf{Performance}_{student} = \{\{3Pass, Fail\}, \{Pass, 2Fail\}\} = 7 \text{ outcomes.}$$

Therefore, in the set $A_1 = Select$,

$$Pr(Fail_{SOB}) \equiv Pr(SOB) = \frac{1}{7}$$

Similarly, in the set $A_2 = Delete$,

$$Pr(Fail_{DS,DW}) \equiv Pr(Fail_{Delete}) = \frac{2}{7}$$

From *e.q.18*,

$$Pr(Fail_{SOB}|A) = \frac{\frac{1}{2} * \frac{1}{7}}{\left(\frac{1}{2} * \frac{1}{7}\right) + \left(\frac{1}{2} * \frac{2}{7}\right)} = 0.33 \cong 33\%.$$

### 3.4. Complement Theory for Prediction of Learning Materials

Complement seeks to sort out the quantity that is missing from a whole sum of events either by difference operation or by negation. In set theory, suppose the set $A = \{a, b, c, d, f\}$ and set $B = \{b, d, f\}$; then $A - B = \{a, c\}$. Expressed as the complement of B, we have

$$\neg B = \{A - B\}$$

and in probability as

$$Pr(A^c) = 1 - Pr(A) \quad (19)$$

Our application of the complement theory in computing learning object or material recommendation requires that the lengths of the elements of set $A$ *(i.e. minuend)* and that of set $B$ *(subtrend)* in the difference operation be equal in number of elements e.g. the set $A = \{PPPP\}$ and set $B = \{PPFP\}$.

### 3.5. Definition

Given that the sets $A$ and $B$ have equal length of elements, the complement of probability between the sets $A$ and $B$ is

$$Pr(A) - Pr(B) = Pr(\{x \mid x \in B \text{ and } x \neg \in A\}) \quad (20)$$
$$Pr(\{PPPP\}) - Pr(\{PPFP\}) = Pr(\{F\})$$

where the weighted probability of **Pass**

$$Pr(\{PPPP\}) \equiv \sum_{i=1}^{n} Pr(P_i) \equiv argMaxPr(PPPP) = 1.$$

Thus, the difference of probability of the set $A$ and set $B$ is

$$argMaxPr(\{PPPP\}) - Pr(\{PPFP\}) = Pr(\{F\})$$

which is the probability of the *failed* learning object, and by computation is

$$1 - \frac{3}{4} = \frac{1}{4} \cong 0.25$$

As a model of computation for pre-assessment and material recommendation in learning and teaching, in Figure 3 is a **corollary** for a generalized probabilistic approach in this work.

Corollary 1:

> In a finite set of $n$ learning objects where $1 \leq i \leq j \leq k \leq n$ such that each object $i$ have both the *Pass* and *Fail* decision states for student pre-learning assessment skillset status; if a given student is pre-assessed on some $n$ learning objects, then the difference of probability between the *Maximum Expected Probability* of the $i$ objects and the *probability* of $j$ number of *Passes* earned by the student is equal to the *probability* of the $k$ number of *failed* learning object probability. That is,
> $$argMaxPr(Pass_1 \ldots Pass_{i+1} | A) - Pr(Pass_1 \ldots Pass_{j+1} | A) = Pr(Fail_1 \ldots Fail_{k+1} | A)$$
> Otherwise;
> $$argMaxPr(Pass_1 \ldots Pass_{i+1} | A) - Pr(Pass_1 \ldots Pass_{j+1} | A) = 0$$
> and
> $$argMaxPr(Pass_1 \ldots Pass_{i+1} | A) - Pr(Fail_1 \ldots Fail_{j+1} | A) = 1$$

Figure 3. Difference of probability generalization for prediction of learning object

The second part of the Corollary equates to 0 (Fig. 3). In information theory, if entropy H(s) = 0, then there is no *impurity* in the given attributes of a dataset. In analogy,

$$argMaxPr(Pass_1 \ldots Pass_{i+1} | A) - Pr(Pass_1 \ldots Pass_{j+1} | A) = 0$$

implies that a student *lacks no skills* amongst the skillsets pre-assessed upon: The interpretation of the expression equals to zero means there is *no impurity* in the student "*knowledge*" within the range of the pre-assessed learning objects. Conversely,

$$argMaxPr(Pass_1 \ldots Pass_{i+1} | A) - Pr(Fail_1 \ldots Fail_{j+1} | A) = 1$$

implies that the student has gained no knowledge, as such will have to relearn all concepts.

## 4. DISCUSSION

Previous and related studies of [21] [23] have described and formalized logical techniques for the pre-assessment of skills and recommendation of learning materials. This covered the use of inductive logic programming and inter-agent (multi-agent) communication system. Based on the extract of the dataset in [23], this paper extends the recommendation of appropriate, relevant and real-time learning materials but in the perspective of probability theorem.

### 4.1. Computing the Probability for Passed Pre-assessments

Firstly, from *eq.11*, we can compute the probability of a student scoring a *pass* given $<Pass | Fail>$ on any leafnode object as

$$Pr(Pass) = \frac{1}{2}.$$

As independent action events, a student can only be in one state i.e. either *Pass* or *Fail* at any given time based on his/her skill sets. The outcome of either a *Pass* or *Fail* state signals the transition to a new state after an ***action*** on a pre-assessment event. The *action* taken is the submission of an answer response to a quiz on a pre-assessment system. The $Pr(Pass)$ is a measure of skill sets that the student has already acquired. We can compute the probability of *skill competencies* using complement theory to determine:

International Journal of Artificial Intelligence and Applications (IJAIA), Vol.13, No.5, September 2022- *The knowledge already acquired that will propel the student to a higher learning;* or
- *The knowledge required to be learned before any progression to a higher-level learning.*

**4.2. Complement Theory for Leafnode $N_i$ Recommendation: $N_i \subsetneq C$ Parent-Node**

Let $N_i$ be some set of leafnodes underneath a prerequisite parent node C [21] [22] [23]. Computing a **Pass** probability over some learning objects asserts that a student has passed the learning object pre-assessed upon and that the student has acquired some (but not all) skills. On the other hand, the recommendation for a higher learning object after pre-assessment requires that **all** the learning objects pre-assessed upon must be *passed*. Otherwise, in this paper we show that if some learning objects are *failed*, then the complement of **Pass** probability $Pr(\neg Pass_j)$ computation be used in the computation of weighted-probability of the needed or requisite knowledge.

Now consider the set of student performance on a single parent node using *eq.16*. Based on the first part of our *Corollary* (Fig. 3) or *eq.19*,

$$argMaxPr(Pass_1, \ldots, Pass_i \mid A) - Pr(Pass_1, \ldots, Pass_j \mid A) = Pr(Fail_1, \ldots, Fail_k \mid A)$$

the probability of a **failed** learning recommendation given the node

$$Select = \{Fail, Pass, Pass, Pass\}$$
$$Pr(PPPP \mid A = Select) - Pr(PPPF \mid A = Select) = Pr(F \mid A = Select)$$
$$= 1 - \frac{3}{4} = \frac{1}{4} = 0.25$$

and for the set of *Delete* where the student performance $A = \{Pass, Fail, Fail\}$, the probability of the recommendation of *failed* learning recommendation is also computed as

$$Pr(PPP \mid A = Delete) - Pr(PFF \mid A = Delete) = Pr(FF \mid A = Delete)$$
$$= 1 - \frac{1}{3} = \frac{2}{3} = 0.67.$$

From the foregoing, we state that

$$[Pr(PPPP \mid A) - Pr(PPPF \mid A)] - [Pr(PPP \mid A) - Pr(PFF \mid A)] = Pr(F \mid A) + Pr(FF \mid A).$$

Recall that our *corollary* also states that

$$argMaxPr(Pass_1 \ldots Pass_i \mid A) - Pr(Pass_1 \ldots Pass_j \mid A) = 0$$

This is the case in which all learning objects $i \in S$ that are pre-assessed upon are **all Passed**: no prerequisite to learn. The difference of probability = 0 implies that in the exercise, a **Failed** pre-learning assessment will never occur; and when this is the case, the student is recommended an immediate higher learning objects of a parent node i.e. the desired learning object (or topic) that invoked the pre-assessed objects. For example, if the pre-assessed parent object $A = Delete$, then the desired topic will be $DesiredConcept = Update$ (Fig. 1). Therefore, it holds that

$$probabilityFail_k = \frac{(argMaxProbability - numberPass_j)}{\sum_{i=1}^{n} argMaxProbability - numberPass_j}$$

and the algorithm given in Figure 4 and the architecture in Figure 5.

Algorithm 1. Probability for Failed Learning Object Recommendation

1. Start
2. argMaxProbability ← 1
3. $a_i = [\ ]$  /** student performance string array */
4. $a_i \leftarrow length of the set a_i; 1 \leq i \leq j \leq k \leq n$
5. $numberPass_j \leftarrow length of Passes in a_i$
6. $probabilityFail_k = \frac{(argMaxProbability - numberPass_j)}{\sum_{i=1}^{n} argMaxProbability - numberPass_j}$
7. Print $probabilityFail_k$
8. $recommendation = \begin{cases} learn\ k\ object\ if\ probabilityFail_k \leq 1 \\ learn\ n+1\ object\ if\ probabilityFail_k = 0 \end{cases}$
8. End

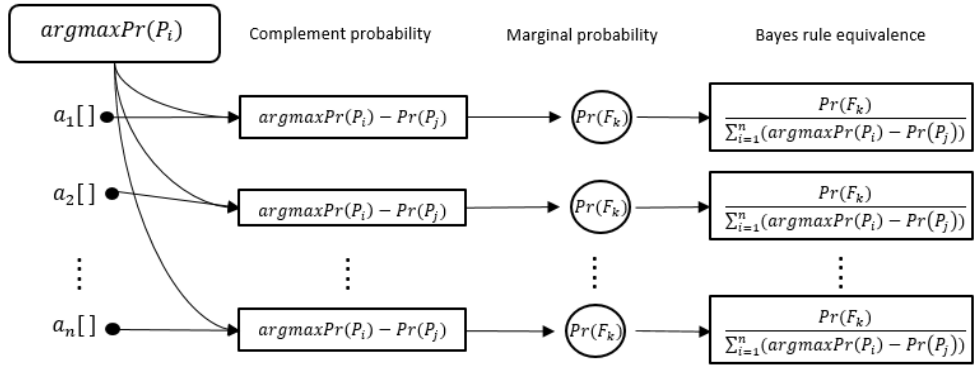

Figure 4. Architecture of Complement of Probability: Difference probability.

## 4.3. Iterative Computation of Difference of Probability Operation

Every assessment has a number of *i* learning objects whose *maximum expected value = 1*. As earlier stated e.g. in *eq.19*; the difference probability determines the weight of the learning material recommended such that 0 implies maximum knowledge with no impurity in the student skillset, and 1 being the worst-case scenario where the student has not gained any competency on the pre-assessed number of *i* learning objects. Therefore, given the set of *i* learning objects where $1 \leq i \leq 3$, we can iteratively determine the weighted probability of the recommended material using the difference probability operation as follows:

If $i = 1$ then $argmaxPr(P_1) = 1$:
  $argmaxPr(P_1 | A) - Pr(F_1|A) = 1$        *To learn 100% Failed object.*
  $argmaxPr(P_1 | A) - Pr(P_1|A) = 0$        *To progress higher. No Failed object.*

If $i = 2$ then $argmaxPr(P_1P_2) = 1$:
  $argmaxPr(P_1P_2 | A) - Pr(P_1F_2|A) = 0.5$        *To learn 0.5 or 50% Failed object.*
  $argmaxPr(P_1P_2 | A) - Pr(F_1PF_2|A) = 0.5$        *To learn 0.5 or 50% Failed object.*
  $argmaxPr(P_1P_2 | A) - Pr(F_1F_2|A) = 1$        *To learn 100% Failed object.*
  $argmaxPr(P_1P_2 | A) - Pr(P_1P_2|A) = 0$        *To progress higher. No Failed object.*

If $i = 3$ then $argmaxPr(P_1P_2P_3) = 1$:

$argmaxPr(P_1P_2P_3 | A) - Pr(P_1P_2F_3|A) = 0.33$  *To learn 0.33 or 33% Failed object.*

$argmaxPr(P_1P_2P_3 | A) - Pr(P_1F_2P_3|A) = 0.33$  *To learn 0.33 or 33% Failed object.*

$argmaxPr(P_1P_2P_3 | A) - Pr(F_1P_2P_3|A) = 0.33$  *To learn 0.33 or 33% Failed object.*

$argmaxPr(P_1P_2P_3 | A) - Pr(P_1F_2F_3|A) = 0.67$  *To learn 0.67 or 67% Failed object.*

$argmaxPr(P_1P_2P_3 | A) - Pr(F_1P_2F_3|A) = 0.67$  *To learn 0.67 or 67% Failed object.*

$argmaxPr(P_1P_2P_3 | A) - Pr(F_1F_2P_3|A) = 0.67$  *To learn 0.67 or 67% Failed object.*

$argmaxPr(P_1P_2P_3 | A) - Pr(F_1F_2F_3|A) = 1$  *To learn 100% Failed object.*

$argmaxPr(P_1P_2P_3 | A) - Pr(P_1P_2P_3|A) = 0$  *To progress higher. No Failed object.*

If $i = \infty$ then ...

For a given number of learning objects $n$, Table 3 illustrates further the probability computation and in Figure 5 the probability distribution of the ratio of **Pass** probabilities $P_i$ to that of **Fail** probabilities $F_i$ over a given sample size $n$. As shown in Table 3, the computed values depict weighted learning materials. For instance, for $n = 5$; if $P5 = 0.2$, then $F5 = 0.8$: an indication that 0.2 is the weighted learning material that was passed and 0.8 the weighted-learning material failed. Thus, computation of weighted-probability learning material for $P_i$ can be determined from the computation of weighted probability of $F_i$ and vice versa; such that, if $\delta \in \{P, F\}$ and $0 \leq \delta \leq 1$, then $Pr(P_i) + Pr(F_i) = 1$.

Table 3: Probability of $\delta \in \{P_i, F_i\}$ for $0 \leq \delta \leq 1$ and n = 1, 2, 3, ..., n

| *n* Size Objects | *P1* | *F1* | *P2* | *F2* | *P3* | *F3* | *P4* | *F4* | *P5* | *F5* | *P6* | *F6* | *P7* | *F7* | *P8* | *F8* |
|---|---|---|---|---|---|---|---|---|---|---|---|---|---|---|---|---|
| 1 | 1 | 0 | 0 | 1 | | | | | | | | | | | | |
| 2 | 1 | 0 | 0.5 | 0.5 | 0 | 1 | | | | | | | | | | |
| 3 | 1 | 0 | 0.67 | 0.33 | 0.33 | 0.67 | 0 | 1 | | | | | | | | |
| 4 | 1 | 0 | 0.75 | 0.25 | 0.5 | 0.5 | 0.25 | 0.75 | 0 | 1 | | | | | | |
| 5 | 1 | 0 | 0.8 | 0.2 | 0.6 | 0.4 | 0.4 | 0.6 | 0.2 | 0.8 | 0 | 1 | | | | |
| 6 | 1 | 0 | 0.83 | 0.17 | 0.67 | 0.33 | 0.5 | 0.5 | 0.33 | 0.67 | 0.17 | 0.83 | 0 | 1 | | |
| 7 | 1 | 0 | 0.86 | 0.14 | 0.71 | 0.29 | 0.57 | 0.43 | 0.43 | 0.57 | 0.29 | 0.71 | 0.14 | 0.86 | 0 | 1 |

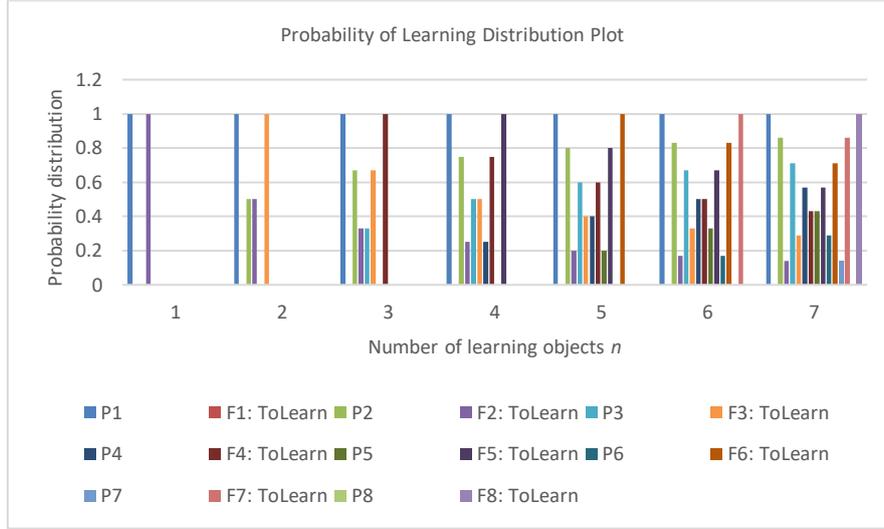

Figure 5. The weighted-probability distribution plot for recommendation of learning. Only column P1 (from Table 3) across all learning objects depict **$Passes$** for all prerequisite pre-assessments and subsequently the recommendation for students' $Desired\_Concept$ $D$. While other columns indicate the recommendation of **$Failed$** learning of **some** or **all** prerequisite leaf nodes $N_i$.

## 4.4. Dataset and Learning Material Prediction by Bayes' Rule

The preceding sections have demonstrated two theoretical approaches, namely, difference of probability over passed and failed learning objects and Bayes' theorem for the prediction and recommendation of learning materials in teaching and learning process. In this section, we present weighted-probability computation based on the student pre-assessments' dataset [23]. In the dataset there are *parent node* $C_i$ and *leafnode* $N_i$ objects (Table 4); and as shown in $eq. 8$ and 9, respectively, $C_i$ is the set that contains the elements $N_i$. In this computation of learning object prediction using Bayes' rule, we consider two parent node objects i.e. $C = \{Select, Delete\}$ over a number of leafnode elements $N_i$.

Table 4: An SQL pre-assessment dataset showing student performance [23]

| Parent-node object | Leafnode object | Pass | Fail | Total |
|---|---|---|---|---|
| Select | SelectOrderBy[SOB] | 11 | 3 | 14 |
| | SelectDistinct[SD] | 14 | 1 | 15 |
| | SelectWhere[SW] | 14 | 0 | 14 |
| | SelectAll[SA] | 14 | 0 | 14 |
| Delete | DeleteSelect[DS] | 5 | 16 | 21 |
| | DeleteWhere[DW] | 20 | 1 | 21 |
| | TruncateTable[TT] | 10 | 3 | 13 |
| Total | | 88 | 24 | 112 |

From the data (Table 4), any leafnode that is failed is recommended for re-learning. Note that the number of **Pass** and **Fail** represents students' performance for each respective leafnode element. In the hierarchy of nodes in our ontology tree (Fig. 1), a parent node $c \in C$ has a higher height than its leaftnodes $N_i$. Now, consider the leafnode $DS$ that recorded 16 Fails. We compute the weighted probability of having the $DS$ node recommended (from the set $Delete$) relative to other nodes that are recommended. That is, the event of choosing a Desired_Concept or topic (parent

International Journal of Artificial Intelligence and Applications (IJAIA), Vol.13, No.5, September 2022node) is prior to the event of pre-assessments on the leafnodes. Thus, in the pre-assessment system, a student chooses a topic ($Desired\_Concept$ denoted as $D$) to study prior to pre-assessments on the prerequisite leafnodes $C \supseteq N_i$ underneath the chosen topic $D$. From $eq.18$, according to Bayes', we have

$$Pr(Fail_{DS}|A = Delete) = \frac{\Pr(Delete \cap Fail_{DS})}{\Pr(Select \cap Fail_{Select)} + pr(Delete \cap Fail_{Delete})}$$

$$= \frac{\frac{55}{112} * \frac{16}{20}}{\left(\frac{57}{112} * \frac{4}{24}\right) + \left(\frac{55}{112} * \frac{20}{24}\right)} = \frac{0.49 * 0.8}{0.51 * 0.17 + 0.49 * 0.83} = 0.78 \cong 78\%.$$

From the foregoing; our customized model predicts that 78% of pre-assessments on the $DS$ (*DeleteSelect*) leafnode would be recommended to study to re-learn the $DS$ query for mastery.

### 4.5. Information Entropy on Student Cognitive Activities

*Information gain (IG)* and *Entropy H(s)* are important metrics in the construction of J48 decision trees. While $IG$ was used to determine the most important attribute (**Select, Insert, Delete, Update, Join**) that would become the root node in the decision tree, the $H(S)$ computes the amount of **impurity** present in each of the **features** on every given attribute in the dataset. In this work, the computed $H(S)$ in Table 6 depicts the values that quantifies the amount of skill-sets information that is present in the data.

Table 5. Pre-assessment Dataset of student performance: extracted from [23]

| Pre-assessment Dataset | | | | | |
|---|---|---|---|---|---|
| **Select** | **Insert** | **Delete** | **Update** | **Join** | **Outcome** |
| SelectOrderBy | InsertInto | DeleteWhere | UpdateSelect | InnerJoin | Fail |
| SelectDistinct | InsertSelect | DeleteWhere | UpdateWhere | FullOuterJoin | Pass |
| SelectWhere | InsertInto | DelectSelect | UpdateSelect | SelectJoin | Pass |
| SelectDistinct | InsertInto | DeleteWhere | UpdateWhere | InnerJoin | Pass |
| SelectDistinct | InsertInto | DeleteSelect | UpdateSelect | InnerJoin | Fail |
| SelectAll | InsertSelect | DeleteSelect | UpdateWhere | InnerJoin | Pass |
| SelectWhere | InsertSelect | DeleteSelect | UpdateWhere | FullOuterJoin | Pass |
| SelectAll | InsertInto | DeleteSelect | UpdateWhere | FullOuterJoin | Pass |
| SelectOrderBy | InsertSelect | DeleteSelect | UpdateSelect | FullOuterJoin | Fail |

In Table 5, we have the Outcome column with 6 Passes, and 3 Fails. Therefore, from

$$H(S) = -\sum_{x \in N} P(x) log_2 P(x)$$

the *Entropy(Outcome)* ≡ *Entropy(6, 3)* i.e. passes = 6, fails = 3; where the number of passes and fails equals 9. Thus, the Entropy before the decision tree split is

$$= -\frac{6}{9} log_2 \frac{6}{9} - \frac{3}{9} log_2 \frac{3}{9}$$
$$= -0.667 * -0.584 - 0.333 * -1.586$$
$$= 0.39 + 0.528$$
$$= 0.918 \text{ uncertainty.}$$

International Journal of Artificial Intelligence and Applications (IJAIA), Vol.13, No.5, September 2022

Subsequently, the information gain given the formula

$$IG(S,A) = H(S) - \sum_{i=0}^{n} P(x)log_2 P(x)$$

before and after the split for each attribute, as computed, is presented in Table 5.

### 4.5.1. Zero Impurity

In information theory, entropy $H(S) = 0$ implies zero impurity. Thus, in this paper, we interpret $H(S) = 0$ to be very high information i.e. there is no impurity in student knowledge of the given learning object or subject area. Whereas, those with high $H(S) = 1$ or close to 1 implies little or no knowledge of the given tasks. As shown in Table 6, the leafnodes $SOB$, $SW$, $SA$, $UW$, and $SJ$ have $H(S) = 0$ impurity. Recall that in the preceding sessions we have demonstrated how *complement of probability eq.* 19, *difference of probability eq.* 20, and *Bayes' rule eq.* 12 – 15 can be applied to predict the probable weight (through probability computations) of learning materials given the performance of students on the basis of the nodes in our ontology tree (Fig. 1). In comparing Bayes probability to the $H(S) = 0$ impurity, now consider the customized Bayes' $eq.15$: suppose there is no **Fail** in the number of leafnodes that were considered, then $Pr(Fail_j|A) = 0$. Since there was no record of Fail, then this is invariably the probability of the recommendation of the student chosen topic called the $DesiredConcept$ $D$ in [21, 22, 24]. Thus, for teaching and learning, we draw the conclusion that where there is 0 impurity, then students' knowledge and competency in these learning units are very high. While for instance, the $H(S)$ of the leafnode $SD$ is computed as 0.306 or 31% lack of skill-set in the knowledge task; that of the leafnode $DW$ resulted in 0.612 or 61% higher-lack of knowledge information. Therefore, from the notation $0 \leq H(S) \leq 1$; we state that if information entropy $H(S) = 0$, then there is no lack of skill-set by the student on a given leafnode; otherwise some skill-set are needed to be filled for mastery of the given learning object.

Table 6. Information gain pre-assessment dataset and entropy of performance attributes.

| Info Gain G(S, A) | | Entropy H(S) | | Info Gain G(S, A) | | Entropy H(S) | |
|---|---|---|---|---|---|---|---|
| Select | 1.219 | SOB | 0 Impurity | Update | 0.558 | US | 0.36 Impurity |
| | | SD | 0.306 Impurity | | | UW | 0 Impurity |
| | | SW | 0 Impurity | Join | 0.834 | SJ | 0 Impurity |
| | | SA | 0 Impurity | | | FOJ | 0.36 Impurity |
| Insert | 0.738 | IS | 0.54 Impurity | | | IJ | 0.444 Impurity |
| | | II | 0.306 Impurity | | | | |
| Delete | 1.225 | DS | 0.306 Impurity | | | | |
| | | DW | 0.612 Impurity | | | | |

## 4.6. Decision Trees: the J48 Algorithm in Weka

Decision tree is a binary classification algorithm where the attributes in a dataset are split continuously so as to reach decision states or labels. In Table 4 are the dataset that contains two class labels **Pass** and **Fail**, five attributes and their respective features. As stated in [25], J48 performs accurate results for classification problem. Thus, J48 [26] algorithm was chosen to implement the decision tree classification model given the set of attributes, features and class labels (Figure 6). The decision tree is a pruned tree in which the *Update* node splits into the *UpdateSelect* and *UpdateWhere* features and, subsequently, categorized into the **Pass** and **Fail** labels. The computed values for information gain $(IG)$ and information entropy $H(S)$ of the skill-set pre-assessment data are presented in Table 6. From Table 6, the *Delete* node has the highest $IG = 1.225$ as well as the highest $H(S) = 0.612$ for $DW$; and *Update* with the lowest $IG = 0.558$. The decision tree in Figure 6 is built based on the best attribute and feature metrics of $IG$ and $H(S)$, respectively: that is, the Update node has the lowest impurity in the dataset. The decision tree depicts the *Update* attribute and its *UpdateSelect* feature with 3 correctly classified instances of **Fail** and 1 incorrectly classified instance of **Pass**, and the *UpdateWhere* feature with all 5 instances correctly classified as **Pass**. This means that, based on the pre-assessment dataset, the *UpdateSelect* learning object is predicted to be recommended for re-learning going by the number of 3 **Fail** instances to 1 **Pass** instance in the decision tree. Of the 9 total number of instances, 8 were correctly classified with 1 misclassification. However, in comparison to *RandomForest*, the confusion matrix showed no instance was incorrectly classified. Thus RandomForest performs better in classification than J48.

## 4.7. Implication of Misclassification of Learning Objects to Student Learning

We take special interest in the **1** *incorrectly* classified learning instance by the J48 decision tree and discusses the implication of this on a given student. Whereas **4 Fail** instances are correctly classified and recommended to be relearned (Fig. 6); the value 1 which is a misclassification is predicted to be a **Pass**. The implication of this *1 out of 5* classification is that learning has occurred in the student, whereas it has not. From Figure 6, the implication is that a given student should skip the learning of the *failed* unit of learning instead of re-learning and re-assessment. This should not be the case for a recommendation system in teaching and learning. Thus, for classification in teaching and learning, 100% accurate classification is unavoidable. Avenues for misclassification must be avoided otherwise gaps of learning would be left in the students. On the basis of the dataset, this decision tree algorithm, has left a gap in student's skill set even with 80% training and 20% test data.

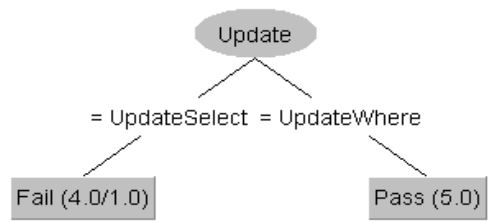

Fig. 6. J48 pruned decision tree visualization. The *Update* node has the least $IG(S, Update) = 0.558$ that makes the most important decision tree with reduced impurity.

In teaching-learning systems, misclassification of learning objects for recommendation is not in the best interest of a student: if some unit of learning that has not been learned is categorized to have been learned as shown by the decision tree classifier. Thus, teaching and learning algorithm such as the inductive logic programming (ILP) as applied in [21] [24] must be used to deliver accurate and exact materials for student learning. This is because learning internalizes structures of concepts and contents that, *1) meets the set standards of institutions*, *2) gets institutional evaluation*, and *3) post-learning application of knowledge to meet societal expectations*; as against shopping-item recommendation systems that could be: 1) based on selective choice of the individual, 2) incidental, and 3) frivolous adventure. Thus, shopping-item recommendation is different from learning-material recommendation.

## 5. CONCLUSIONS AND FURTHER WORK

Learning is an intensive task in which high cognitive functions are required on the part of the student. This paper has demonstrated different computational models, namely; complement of probability, difference of probability, Bayes' rule, and information entropy in the computation and prediction of learning materials to students. We have demonstrated that when $Pr(\neg Fail_i) = 0$, then it is the recommendation for an immediate higher-level learning; then for $0 < Pr(\neg Pass_i) \leq 1$, it is the recommendation for the weighted k-number of **Failed** learning object. We showed that the complement/difference of probability and Bayes' rule are equally applicable in the prediction of accurate learning for students. We have used J48 ML algorithm to show misclassification of feature in ML –- which is not suitable for the mastery of learning content. For the avoidance of leaving gaps in students' knowledge, computational models must be 100% accurate in skills classification problem. In order for LTA (learning, teaching and assessment) systems to support mastery of subject matter content, computational models and ML algorithms that supports pre-learning assessment process should be integrated into LTA to have Pre-assessment LTA (PLTA) system. In future, this project shall look at moving this application to a web-based learning environment and implementing learning recommendation based on inductive logic programming and/or using the complement of probability and our customized Bayes' rule.

## STATEMENTS AND DECLARATIONS

There are non-financial interests that are directly or indirectly related to the work submitted for publication.

## CONFLICT OF INTEREST

The authors declare that they have no conflict of interest.

## REFERENCES

[1] E. K. Ehimwenma, S. Krishnamoorthy and S. Al Sharji, "Applied Complement of Probability and Information Entropy for Prediction in Student Learning (Abstract)," World Academy of Science, Engineering & Technology (WASET); International J. Computer and Information Engineering, vol. 15, no. 3, 2021.

[2] D. Saad, "Information Theory, Inference, and Learning Algorithms," American Scientist, vol. 92, no. 6, p. 578, 2004.

[3] G. Bonanno, Game Theory, 2nd ed., California: CreateSpace Independent, 2018, pp. 1-592.

[4] S. Lipschutz and J. J. Schiller, Introduction to Probability and Statistics, USA: The McGraw-Hill Companies, 2012, pp. 1-370.

[5] C. Batanero, E. J. Chernoff, J. Engel, H. S. Lee and E. Sánchez, "Research on teaching and learning probability," Springer Nature, Vols. ICME-13 Topical Survey, pp. 1-40, 2016.

[6] M. Aldridge, O. Johnson and J. Scarlett, " Group testing: an information theory perspective.," arXiv preprint arXiv:1902.06002., 2019.

[7] J. Zhang, "A study on mental health assessments of college students based on triangular fuzzy function and entropy weight method," Mathematical Problems in Engineering,, 2021.

[8] X. B. Liu, Y. J. Zhang, W. K. Cui, L. T. Wang and J. M. Zhu, "Development assessment of higher education system based on TOPSIS-entropy, hopfield neural network, and cobweb model," Comlplexity, 2021.

[9] R. Joshi and M. Alehegn, "Analysis and prediction of diabetes diseases using machine learning algorithm: Ensemble approach," International Research Journal of Engineering and Technology, vol. 4, no. 10, pp. 426-435, 2017.

[10] N. C. D. Adhikari, "Prevention of heart problem using artificial intelligence," International Journal of Artificial Intelligence and Applications (IJAIA), vol. 9, no. 2, 2018.

[11] S. S. Cheeti, "Twitter based sentiment analysis of impact of covid-19 on education globaly," International Journal of Artificial Intelligence and Applications (IJAIA), vol. 12, no. 3, 2021.

[12] D. Sardana, S. Marwaha and R. Bhatnagar, "Supervised and Unsupervised Machine Learning Methodologies for Crime Pattern Analysis," International Journal of Artificial Intelligence and Applications (IJAIA), vol. 12, no. 1, 2021.

[13] S. T. Gopalakrishna and V. Vijayaraghavan, "Automated tool for Resume classification using Sementic analysis," International Journal of Artificial Intelligence and Applications (IJAIA), vol. 10, no. 1, 2019.

[14] J. D. Rennie, L. Shih, J. Teevan and D. R. Karger, "Tackling the poor assumptions of naive bayes text classifiers," in 20th International Conference on Machine Mearning (ICML-03) (pp. 616-623), 2003.

[15] P. Domingos and M. Pazzani, "Beyond independence: Conditions for the optimality of the simple bayesian classifier," in 13th Intl. Conf. Machine Learning (pp. 105-112), 1996.

[16] B. Seref and E. Bostanci, "Performance of Naïve and Complement Naïve Bayes Algorithms Based on Accuracy, Precision and Recall Performance Evaluation Criterions," International Journal of Computing, vol. 8, no. 5, pp. 75-92., 2019.

[17] A. Ashari, I. Paryudi and A. M. Tjoa, "Performance comparison between Naïve Bayes, decision tree and k-nearest neighbor in searching alternative design in an energy simulation tool," International Journal of Advanced Computer Science and Applications (IJACSA), vol. 4, no. 11, 2013.

[18] R. Nithya, D. Ramyachitra and P. Manikandan, "An Efficient Bayes Classifiers Algorithm on 10-fold Cross Validation for Heart Disease Dataset," International Journal of Computational Intelligence and Informatics, vol. 5, no. 3, 2015.

[19] A. Goyal and R. Mehta, "Performance Comparison of Naïve Bayes and J48 Classification Algorithms," International Journal of Applied Engineering Research, vol. 7, no. 11, 2012.

[20] C. M. Grinstead and J. L. Snell, Grinstead and Snell's Introduction to Probability: The CHANCE Project, American Mathematical Society, 2006.

[21] K. E. Ehimwenma, P. Crowther and M. Beer, "Formalizing logic based rules for skills classification and recommendation of learning materials," Int. J. Inf. Technol. Comput. Sci.(IJITCS), vol. 10, no. 9, pp. 1-12, 2018.

[22] K. E. Ehimwenma, P. Crowther, M. Beer and S. Al-Sharji, " An SQL Domain Ontology Learning for Analyzing Hierarchies of Structures in Pre-Learning Assessment Agents," SN Computer Science, vol. 1, no. 6, pp. 1-19, 2020.

[23] K. E. Ehimwenma and S. krishnamoorthy, "Design and Analysis of a Multi-Agent E-Learning System Using Prometheus Design Tool," International Journal of Artificial Intelligence (IJ-AI) SCOPUS, vol. 9, no. 4, pp. 31-45, 2020.

[24] K. E. Ehimwenma, "Chapter 4: Methodology: Agent Oriented Analysis & Design and Classification," in A multi-agent approach to adaptive learning using a structured ontology classification system (Doctoral dissertation), Sheffield, Sheffield Hallam University, 2017, pp. 65-102.

[25] N. Saravanan and V. Gayathri, "Performance and classification evaluation of J48 algorithm and Kendall's based J48 algorithm (KNJ48)," International Journal of Computer Trends and Technology (IJCTT), vol. 59, no. 2, pp. 73-80, 2018.

[26] U. o. W. UoW Machine Learning Group, "WEKA The workbench for machine learning," [Online]. Available: https://www.cs.waikato.ac.nz/ml/weka/index.html. [Accessed 02 February 2021].